\documentclass[letterpaper, 10 pt, conference]{ieeeconf}  

\IEEEoverridecommandlockouts                              
\usepackage{graphics} 
\usepackage{graphicx}
\usepackage{xcolor}
\usepackage{amsmath}
\usepackage{mathtools}
\usepackage{cite}
\usepackage{caption}
\usepackage{subcaption}
\usepackage{algorithm}
\usepackage{algpseudocode}
\usepackage{comment}

\title{\LARGE \bf
Adaptive Negative Damping Control for User-Dependent\\ Multi-Terrain Walking Assistance with a Hip Exoskeleton
}

\author{Giulia Ramella$^{1,2}$, Auke Ijspeert$^2$, and Mohamed Bouri$^{1,3}$
\thanks{Giulia Ramella received funding from the European Union's Horizon 2020 research and innovation program under the Marie-Sklodowska-Curie Grant Agreement No. 945363.}
\thanks{$^1$ Giulia Ramella and Mohamed Bouri are with the REHAssist group, Ecole Polytechnique Federale de Lausanne (EPFL). 
        {\tt\small giulia.ramella@epfl.ch}}%
\thanks{$^2$ Giulia Ramella and Auke Ijspeert are with the BioRobotics Laboratory, Ecole Polytechnique Federale de Lausanne (EPFL).}%
\thanks{$^3$ Mohamed Bouri is with the TNE Laboratory, Institute of Neuro-X, Ecole Polytechnique Federale de Lausanne (EPFL).}%
}

\begin{document}
\maketitle
\thispagestyle{empty}
\pagestyle{empty}

\begin{abstract}
Hip exoskeletons are known for their versatility in assisting users across varied scenarios. However, current assistive strategies often lack the flexibility to accommodate for individual walking patterns and adapt to diverse locomotion environments. In this work, we present a novel control strategy that adapts the mechanical impedance of the human-exoskeleton system. We design the hip assistive torques as an adaptive virtual negative damping, which is able to inject energy into the system while allowing the users to remain in control and contribute voluntarily to the movements. Experiments with five healthy subjects demonstrate that our controller reduces the metabolic cost of walking compared to free walking (average reduction of $7.2 \%$), and it preserves the lower-limbs kinematics. Additionally, our method achieves minimal power losses from the exoskeleton across the entire gait cycle (less than $2 \%$ negative mechanical power out of the total power), ensuring synchronized action with the users' movements. Moreover,  we use Bayesian Optimization to adapt the assistance strength and allow for seamless adaptation and transitions across multi-terrain environments. Our strategy achieves efficient power transmission under all conditions. Our approach demonstrates an individualized, adaptable, and straightforward controller for hip exoskeletons, advancing the development of viable, adaptive, and user-dependent control laws.
\end{abstract}

\vspace{+1em}

\section{Introduction}
Hip exoskeletons for walking assistance are gaining popularity for their versatility to assist users across diverse contexts. Initially designed for rehabilitation purposes, these wearable devices are now increasingly used to assist healthy users. In recent years, there have been significant advancements, but developing control strategies that seamlessly synchronize with users remains a major challenge \cite{baud_review_2021}. This is particularly critical with healthy users, where the exoskeleton’s assistive actions must align with voluntary movements and individual intentions. To achieve a synchronized assistance, control algorithms must account for individual gait patterns, terrain variability, and diverse walking styles \cite{slade_personalizing_2022}.

\begin{figure}[t]
\centering
\includegraphics[width=0.9\linewidth]{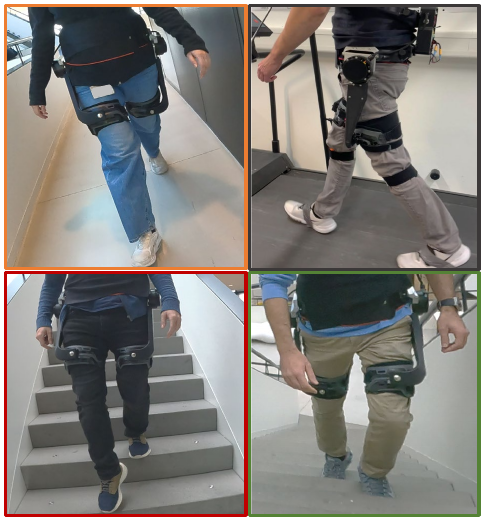}
\caption{The user-dependent and multi-terrain walking assistance with the hip exoskeleton eWalk. Participants walked on treadmill and in varied unstructured environments.}
\label{FirstPage}
\centering
\vspace{-2.5em}
\end{figure}

Multiple control approaches have been proposed to achieve a compliant human-exoskeleton interaction. Adaptive oscillators \cite{seo_fully_2016} and fixed-shape torque profiles \cite{manzoori_evaluation_2024} are effective, but do not address individual walking patterns. Human-in-the-loop optimizations achieve highly personalized control \cite{bryan_optimized_2021}, but they involve computationally-intensive and time-demanding procedures. Energy-shaping methods \cite{zhang_optimal_2023} are precise and effective, but demand complex modeling of the system. 
Previously, researchers used the GEMS exoskeleton to implement a delayed output feedback control that relies on hip joint angular positions \cite{lim_delayed_2019}. However, this approach requires manual tuning of delays, as these signals do not naturally align with hip torques. Similarly, assisting users with estimated biological hip torques is a highly advanced technique but requires tuning of delay parameters \cite{molinaro_estimating_2024}. Overall, further explorations are needed to address the following research question: how can a hip exoskeleton control strategy be designed to align the assistance with the wearer’s voluntary movements while adapting to individual gait patterns across diverse terrains?

When designing control strategies, researchers widely agree on the importance of the positive mechanical power transmitted from the exoskeleton to the user \cite{bryan_optimized_2021, lim_delayed_2019, quinlivan_assistance_2017}. To regulate the energy transfer, existing research has explored a virtual modification of the mechanical impedance of the human-exoskeleton system \cite{aguirre-ollinger_1-dof_2007, aguirre-ollinger_active-impedance_2007, ortlieb_active_2018}. For instance, negative damping has been used to modulate the dynamic response of the human-exoskeleton system, while keeping the wearer in control of the movements \cite{aguirre-ollinger_active-impedance_2007, aguirre-ollinger_1-dof_2007}. Knowing that damping naturally causes energy dissipation, providing negative damping can inject energy into the system. As damping is proportional to the joint velocities, the resulting assistance is inherently dependent on the wearer's motion. 
In theory, negative damping alone introduces positive feedback, risking instability and torque escalation. However, in the practical case of human-exoskeleton interaction, these issues are prevented thanks to pendulum dynamics, musculoskeletal damping, and torque limits. Preliminary implementations with a knee exoskeleton \cite{aguirre-ollinger_1-dof_2007, aguirre-ollinger_active-impedance_2007} suggest that bounded negative damping holds great promise to provide partial assistance with wearable devices. 

In this work, we design a negative damping control for the hip exoskeleton eWalk. With this approach, the exoskeleton torques assist wearers according to their own volitional movements. We experimentally evaluate the impact of our control strategy on users' walking energetics by comparing the assisted and the exoskeleton-free conditions. To assess the synchronization between the users' movements and the exoskeleton torques, we compute the transmitted mechanical powers. This quantity is defined as the product of the exoskeleton torque and the user's hip angular velocity, and positive powers indicate alignment between the users' movements and the assistive torques. Moreover, we measure the lower-limbs kinematics to analyze if the exoskeleton assistance preserves the natural walking movements.

Afterwards, we evaluate our controller's ability to assist users in unstructured environments (Figure \ref{FirstPage}). During multi-terrain locomotion, exoskeletons must dynamically adapt the assistance strength to meet varying demands across terrains \cite{kang_effect_2019, manzoori_adaptive_2024, molinaro_task-agnostic_2024}. Existing studies have proposed terrain recognition algorithms based on the data from onboard sensors \cite{lin_improving_2024, kang_continuous_2020, tricomi_environment-based_2023}, but these methods are prone to misclassification. In contrast, we propose a Bayesian Optimization framework to adjust the assistance level to different scenarios (flat walking and stairs descending/ascending) relying only on the exoskeleton's kinematic signals observed over a defined time window.

In summary, we present a novel control approach that injects energy into the human-exoskeleton system. We design an adaptive virtual negative damping control, where the assistive torques are generated based on the users' hip kinematics. Compared to existing works, our algorithm encompasses both individualization and terrain adaptation, while maintaining a straightforward and versatile approach. The main contributions are: (i) the design of the hip assistive torque profile based on the wearer's volitional movements; (ii) an experimental evaluation of the efficacy of our control to reduce the metabolic cost of walking, while preserving individual walking kinematics; and (iii) a validation of the adaptability of our control across varied locomotion terrains. 

The rest of this paper is organized as follows. In Section II we describe our algorithm for hip exoskeleton assistance, and the methodologies to validate our strategy. Section III presents the results and discussion for the experiments performed on a treadmill and on unstructured environments. Section IV presents the conclusions and the future directions of this study.

\section{Methods}

\subsection{Assistive Control Law}
We designed the torque profile provided by our hip exoskeleton eWalk as a virtual negative damping. This approach is able to inject energy into the system, while keeping the user in control of the movements \cite{aguirre-ollinger_active-impedance_2007, aguirre-ollinger_1-dof_2007}. In order to apply the control action, a model for human locomotion is required. Although human walking has been described by diverse models of varied complexity \cite{faraji_3lp_2016}, we opted for the simple compass-gait biped \cite{spong_passivity-based_2007}, where the legs are represented as rigid links connected at the hip joint. This approach has demonstrated a remarkable ability to capture fundamental characteristics of bipedal locomotion \cite{adamczyk_redirection_2009,
darici_anticipatory_2020, darici_humans_2023, ohtsu_powered_2023}, the legs coupling \cite{kuo_simple_2001}, and the energy exchanges in human-exoskeleton interactions \cite{zhang_optimal_2023}. This method assumes no foot dynamics, and describes walking as a series of alternating swing and stance phases. At each heel-strike event, the stance and swing leg roles are switched between the feet. 

\begin{figure}[t]
\vspace{0.5em}
\includegraphics[width=\linewidth]{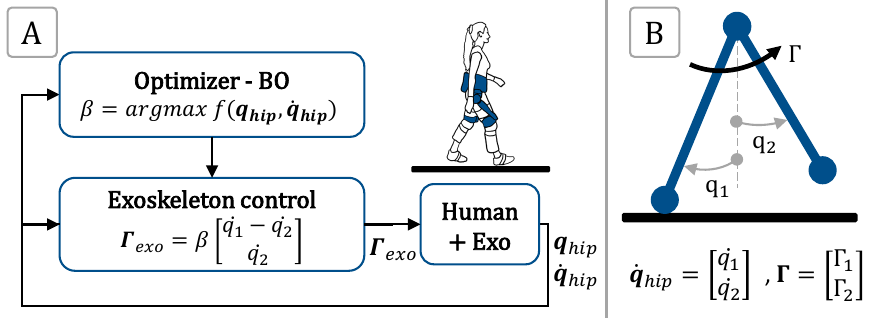}
\caption{(A) Schematic representation of our control framework. The torque profile is defined by a control law consisting of a Virtual Adaptive Negative Damping. The multiplicative coefficient ($\beta$) for the assistance strength is optimized with a Bayesian Optimization (BO). (B) Schematic representation of the compass-gait model and its main variables, where $q_1$ represents the hip angle of the stance leg, and $q_2$ the hip angle of the swing leg.}
\label{algo}
\centering
\vspace{-1.75em}
\end{figure}

The model’s state variables consist of the hip joint positions and velocities for the stance and swing legs, denoted by $q_1$ and $q_2$ respectively ($\mathbf{q}_{hip} =[q_1; q_2]$). The hip input torques are defined as 
$\mathbf{\Gamma} = [\Gamma_1; \Gamma_2] $ 
, as shown in Figure \ref{algo}B. The control is implemented as the product of the coupling matrix $\mathbf{R} =[1 \ -1 \ ; 0 \ 1]$ (inspired from \cite{spong_passivity-based_2007}), and the user's hip joints angular velocity vector $\mathbf{\dot{q}_{hip}}$ measured by the exoskeleton motor encoders at each sampling time. A damping factor $\beta$ ($ Nm s/rad $) allows to modulate the strength of the assistive torques. The resulting the total expression is $\mathbf{\Gamma_{exo}} = \beta \mathbf{R} \mathbf{\dot{q}_{hip}}$ (``Exoskeleton control'' in Figure \ref{algo}A). The value of $\beta$ is bounded to prevent the exoskeleton assistance from disrupting the movements or overpowering the users. The control torque is computed in real time at each sampling step by the exoskeleton's embedded controller.

At heel-strike, the swing leg becomes the new stance leg, causing a reset in the velocities and joint configuration. The coordinate change from before (-) to after (+) the heel-strike event is expressed as: $q^+ = [0 \ 1; 1 \ 0] q^-$. To detect the stance and swing leg, we analyzed the signs of the hip angular velocities as suggested in a previous study \cite{grimmer_stance_2019}. We avoided using joint angular positions to eliminate the ambiguity caused by the surjective mapping of joint angles to gait phases  \cite{baud_review_2021}. Instead, we relied on hip joint angular velocities that can differentiate the stance and swing legs based on their opposite signs during continuous locomotion. We represented the hip flexion with positive increasing angles. The leg with hip extension (negative velocities), and counter-leg in flexion (positive velocities), was identified as $q_1$ (stance), and the other leg as $q_2$ (swing). 

The exoskeleton assistive torques are designed to inject energy into the system, and the resulting total energy of the human-exoskeleton system can be expressed as $E_{tot} = E_{human} + E_{exo}$. By ensuring that $E_{exo}$ remains bounded and lower than the human energy contribution, the exoskeleton assistance synchronizes with the user's walking cycle. Adjusting and limiting the assistive strength ($\beta$) allows the control to dynamically support natural cyclic motion without disrupting the user's gait. 

In order to incorporate the hip angular velocities in the control law, we filtered the signals measured be the exoskeleton motors encoders, as suggested in previous research \cite{lim_delayed_2019, lim_parametric_2023}. Pilot testings were performed to define the cutoff frequency that could reduce the signal peaks while preserving the majority of its frequency components. A power-spectrum analysis guided our choice to filter joint velocities at $4 Hz$. Filtering also ensures that quasi-static movements and static standing (with quasi-zero joint velocity) do not result in a jittery assistance, but keep the torque values close to zero. 


\subsection{Torque Amplitude Optimization}
To modulate the assistance strength across varied locomotion scenario, we implemented a real-time optimization procedure to adapt the multiplicative coefficient $\beta$ (where $\beta \in [1.0,2.5]$, with bounds tuned experimentally). With the Optuna framework \cite{akiba_optuna_2019, bellegarda_quadruped-frog_2024}, we developed a Bayesian Optimization algorithm based on features computed from the hip exoskeleton kinematics (``Optimizer - BO'' in Figure \ref{algo}A). Unlike traditional mapping approaches, this optimization framework enables iterative updates of the algorithm based on the system's status and conflicting objectives. Additionally, designing the optimization based solely on kinematic variables eliminates the need for complex terrain-recognition algorithms or continuous manual tuning of torque coefficients \cite{kang_continuous_2020, grazi_kinematics-based_2022}. 

Our real-time multi-objective optimization balances two objectives. On one hand, we need to increase the torque during larger hip range of motion (ROM, the difference between minimum and maximum measured hip angles), as in stair or slope ascending \cite{montgomery_contributions_2018}. On the other hand, we want to limit the transmitted torque in case of increased movement jerkiness (where the jerk is computed as the derivative of the acceleration). The resulting objective is expressed as $f_{obj}=q_{rom}-\dddot{q}_{mean}$. At the end of a fixed $2 \ s$ observation window, the Bayesian Optimization maximizes the value of the kinematics-based objective function, providing as output the optimal value of $\beta$.

\subsection{Experimental Protocol}
Five healthy subjects volunteered to participate in the study (age: $29 \pm 2.8$ yrs old, height: $1.73 \pm 0.10$ m, weight: $65.8 \pm 6.4$ kg). The experimental procedures were approved by the human research ethics committee of Canton de Vaud (CER-VD) under the protocol number 2023-02305. Participants provided an informed consent before the experimental evaluation, and the study was conducted in compliance with the principles of the Declaration of Helsinki. 

The experimental evaluation was divided into two sessions. Initially, we evaluated the performance of our controller in steady-state walking conditions on a treadmill. Participants were first familiarized with the exoskeleton assistance by walking with the device for 10 minutes on a treadmill. Then, they performed two 5-minute treadmill walking tests, one without the device and one with the exoskeleton assistance. Subjects walked at a fixed speed of $1.1 \ m/s$, and the experimental conditions were proposed in a randomized order. In the assisted condition, the same value of $\beta = 2.3$ was used to facilitate intra-subject comparisons. 

\begin{figure}[t]
\includegraphics[width=\linewidth]{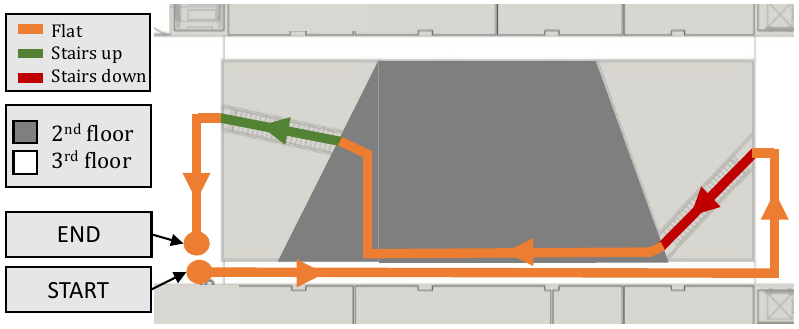}
\caption{Representation of the walking path for the varied terrain locomotion session (top view). The terrain consists of stairs (up and down), flat bouts, and changes of direction.}
\label{methods}
\centering
\vspace{-1.75em}
\end{figure}

In a second session, we investigated the adaptability of our control in an unstructured environment. Participants walked with the assistive device in a varied terrain, consisting of flat sections, stairs ascending/descending, changes in direction, and variations in the walking speed. The duration of each segment was timed and kept consistent across participants, with the total duration set to $90 \ s$. One of the buildings of EPFL was used for this scenario, and the walking path (from a top view) is represented in Figure \ref{methods}.

\vspace{-0.5em}
\subsection{Measurement Setup}
In our study, we used the hip exoskeleton eWalk \cite{manzoori_adaptive_2024}, an active hip orthosis developed in a collaboration between the BioRob-REHAssist group of EPFL and the company Sonceboz. This exoskeleton provides partial assistance during walking, with an actuated joint to assist hip flexion/extension and a passive degree of freedom for hip abduction to facilitate natural walking movements \cite{husty_hibso_2018}. The version of the device used in this study weighs 4.4 $kg$, and it is attached to the user's body with a corset and two thigh attachments. eWalk is actuated by two DC servo motors (PowerSolutions, X8 Pro, maximum torque 32 $Nm$ and reduction ration 1:9, China), and controlled by an embedded computer (BeagleBoneBlack, Texas Instruments, USA) running at 500 $Hz$.

To record the kinematic data of the lower limbs during walking, we used the portable motion capture system XSens (XSens, Movella, The Netherlands). The system has a custom interface to collect and process IMU data and to reconstruct the body movements. Data collection was performed at $100 Hz$, and the IMUs were calibrated for each participant according to the recommended procedure.

We measured the walking energetics in real-time through indirect calorimetry, by using the Q-NRG MAX metabolic monitoring device (COSMED Srl, Rome, Italy). 

During the experiments, we used two separate computers. The first one was used to collect kinematic data in real-time with the XSens software. A second device was connected to the exoskeleton via Wi-Fi in order to monitor the exoskeleton data from a custom GUI. This second computer was also used to perform the optimization during the unstructured walking session.

\vspace{-0.35em}
\subsection{Metrics}
After the experiments, the collected data was processed in custom routines in MATLAB 2023b (The MathWorks, Natick, MA, USA). Before the analysis in MATLAB, XSens data was pre-processed in the custom software in order to reconstruct the walking kinematics of each user. We computed dedicated metrics for all participants, indicated as S1, S2, S3, S4, and S5. For treadmill walking trials, data was collected from the exoskeleton, XSens, and the indirect calorimetry device. For the walking session in the unstructured environment, we recorded data from the exoskeleton.

\subsubsection{Exoskeleton data - treadmill}
From the exoskeleton data, we extracted the applied torque profiles, the values of hip joint angular positions and velocities measured by the motor encoders, and the values from the inertial measurement unit on the exoskeleton corset. These kinematic signals were segmented into individual gait cycles with an algorithm developed in our laboratory \cite{manzoori_gait_2023}. These signals were averaged across gait cycles and between legs to derive the mean profiles for each participant. Finally, we computed the maximum torque value normalized by the weight of each user ($\frac{\Gamma_{exo,max}}{weight}$, in $Nm/kg$).

Power profiles were calculated by multiplying the joint angular velocities ($\omega(t), \ rad/s$) by the torque applied by the exoskeleton ($\Gamma_{exo}(t), \ Nm$), according to the formula $P(t) = \omega(t) \cdot \Gamma_{exo}(t), \text{for } 0 \leq t \leq T_{\text{gait cycle}}$. The resulting power profiles were averaged across gait cycles, and then normalized by each user's weight. We also evaluated the power transmission efficiency between the exoskeleton and the wearer. To this end, we computed the positive power, representing efficient energy transmission, and the negative power, an indicator of energy losses. To quantify power losses, we calculated the percentage of the negative power area relative to the total area under the power profiles.

\subsubsection{Walking energetics data}
To compute the metabolic energy expenditure, we integrated into the Weir equation \cite{weir_new_1949} the values of oxygen and carbon dioxide measured by indirect calorimetry. For each treadmill trial, with and without the exoskeleton, we calculated the mean energy consumption during the final two minutes. Then, we subtracted the resting metabolic rate measured during the two minutes prior to the start of the experiment. This procedure allowed to isolate the activity-related energetic expenditure. Finally, we determined the percentage reduction in metabolic cost (MC) between the two experimental conditions as $\frac{(MC_{exo}-MC_{noexo})}{MC_{noexo}}*100$. 

\subsubsection{Kinematic data}
Lower-limbs kinematic data was extracted from the XSens software after being processed to reconstruct individual movements. We extracted the vertical displacement of the center of mass (COM) and the flexion/extension angles of the hip joint. Then, we segmented the data using the heel-strike information of XSens, and computed the mean values for each subject. We also calculated the cadence ratio between exoskeleton-assisted and device-free walking conditions. Finally, we computed the ratio between the stance time and the swing time, and verified the statistical significance ($\alpha < 0.05$) between the two experimental conditions for each participant.

\subsubsection{Exoskeleton data - unstructured terrain}
When subjects walked in the unstructured environment, we recorded the applied torques, and the hip joint angular positions and velocities measured by the exoskeleton. From this data, we analyzed the torque profiles resulting from our optimization framework across the various walking conditions. To evaluate the efficiency of the power transmission, we calculated the negative power area (representing power losses) as a percentage of the total power area.

\section{Results and Discussion}
In our study, we successfully developed a hip exoskeleton control strategy that assists the wearers based on their volitional movements. The exoskeleton assistance adapts to individual gait patterns, and dynamically adjusts the support across various terrains. In this section, we first discuss the resulting user-specific hip torque assistive profiles. Then, we analyze the effectiveness of the assistive control strategy to reduce the walking energetics, while preserving individual kinematic patterns. Finally, we demonstrate the adaptability of the control framework to assist users in a variety of locomotion terrains.

\begin{figure}[t]
\vspace{1em}
\includegraphics[width=\linewidth]{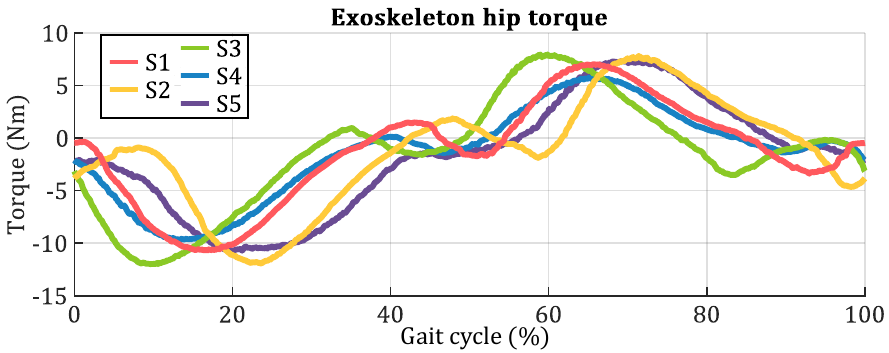}
\caption{Hip exoskeleton torque profiles resulting from our control strategy, for each participant to the experiment, when walking on the treadmill.}
\label{hipTaus}
\centering
\vspace{-2em}
\end{figure}

\subsubsection{Exoskeleton data - treadmill}
We succeeded in generating exoskeleton hip torques to assist users during walking. The exoskeleton torques supported participants during both hip flexion and extension, with smooth transitions between gait cycles. Participants exhibited varied hip assistive torques profiles depending on their individual walking pattern and corresponding joints kinematics (Figure \ref{hipTaus}). The obtained torque profiles closely resemble the exoskeleton hip torques obtained with human-in-the-loop optimization techniques \cite{bryan_optimized_2021}. However, our approach did not require extensive tuning, as the torque curves were derived directly from the users' movements. 

The peak torque timings varied between participants, but they were generally consistent with previous results in the literature \cite{bryan_optimized_2021, lim_parametric_2023, molinaro_task-agnostic_2024,ramella_rapid_2025-2}. Most importantly, these timings were not manually adjusted for each participant. Instead, torque profiles were automatically synchronized with users' movements by incorporating joint angular velocities, thus the movements direction, into the control law. 
This is a crucial aspect because modifying the timing of assistive torques has been proven to affect the total power transferred from the exoskeleton to the user \cite{bryan_optimized_2021, lee_effects_2017, ramella_rapid_2025-2}. 

\begin{figure}[t]
\includegraphics[width=\linewidth]{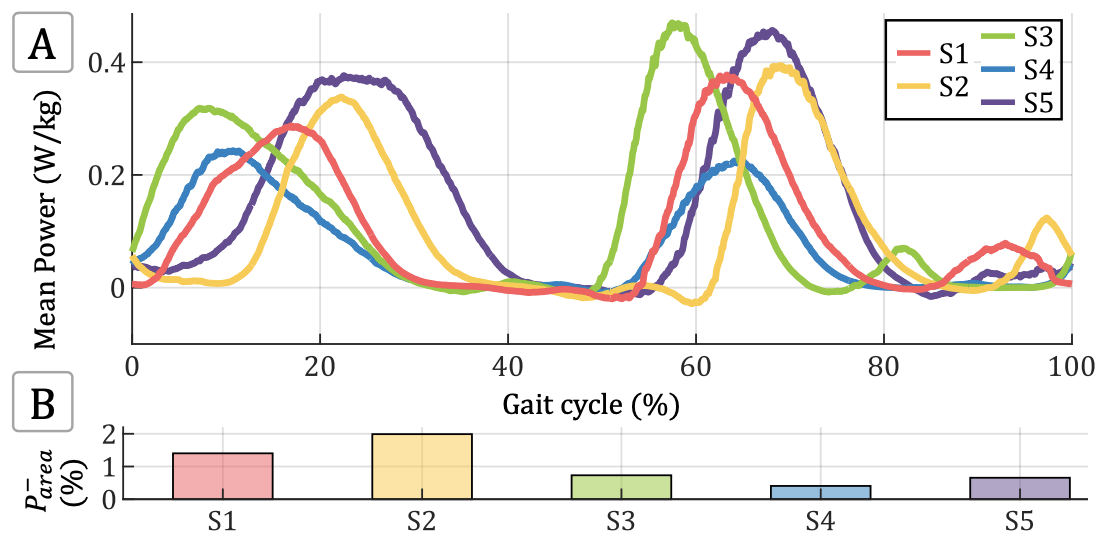}
\caption{(A) Power profiles calculated as the product of the hip exoskeleton torque and the hip joint angular velocities, for each participant. (B) Percentage of negative power area with respect to the total power area.}
\label{powerIN}
\centering
\vspace{-1.5em}
\end{figure}

The power profiles shown in Figure \ref{powerIN}A further emphasize the user-dependent nature of our control strategy. These profiles show peaks at different phases of the gait cycle, reflecting individual movements and torque profiles. Participants showed power profiles with peaks timings similar to the biological powers reported in the literature \cite{reznick_lower-limb_2021}. Conversely, the exoskeleton torque profiles appeared to be delayed compared to their biological counterparts \cite{reznick_lower-limb_2021, winter_human_1995}. 

The power curves remained positive for most of the gait cycle, indicating that our control strategy remained aligned with users' movements. This efficient power transmission is further supported by the percentage of negative power remaining below $2 \%$ for all participants during the gait cycle (Figure \ref{powerIN}B). Negative power indicates torques are not synchronized with the wearers' movements, therefore minimizing this value is essential to ensure seamless human-exoskeleton interaction \cite{ramella_rapid_2025-2, bryan_optimized_2021, lim_delayed_2019, quinlivan_assistance_2017}. 

\begin{table}[tbp]
\vspace{+1em}
\caption{Metrics of our experiment.}
\begin{center}
\vspace{-1em}
\begin{tabular}{c c c c c c}
\textbf{Metric} & \textbf{\textit{S1}} & \textbf{\textit{S2}} & \textbf{\textit{S3}} &
\textbf{\textit{S4}} & \textbf{\textit{S5}}\\ 
\hline
$\frac{(MC_{exo}-MC_{noexo})}{MC_{noexo}}*100 \ (\%)$ & -5.6 & -7.7 & -3.9 & -7.7 & -11.0 \\
$\frac{\Gamma_{exo,max}}{weight}$ \ ($\frac{Nm}{kg}$) & 0.17 & 0.17 & 0.18 & 0.13 & 0.19 \\
$\frac{Cadence_{exo}}{Cadence_{noexo}}$ ( ) & 1.00 & 1.04 & 1.01 & 1.02 & 0.97 \\
\end{tabular}
\label{tab1}
\end{center}
\vspace{-2.75em}
\end{table}

\subsubsection{Walking energetics data}
All participants exhibited reduced metabolic cost when walking with the exoskeleton against the condition without the device (Table \ref{tab1}). In our experiments, we achieved an average reduction of $7.2 \pm 2.6 \%$. This result demonstrates that the energy delivered into the human-exoskeleton system by our negative damping control can lead to a decreased energetic demand for the wearer. 

The variability in the metabolic cost reductions may be attributed to differences in participants' anthropometric characteristics. This is because, despite using the same $\beta$ for all participants, the exoskeleton transmitted the mechanical power differently to each user. Consequently, the shortest participant experienced the greatest decrease in metabolic cost (S5, Table \ref{tab1}). This user also received the greatest power relative to their body weight (Figure \ref{powerIN}), and the highest peak torque normalized by body weight (Table \ref{tab1}).

\begin{figure}[t]
\includegraphics[width=\linewidth]{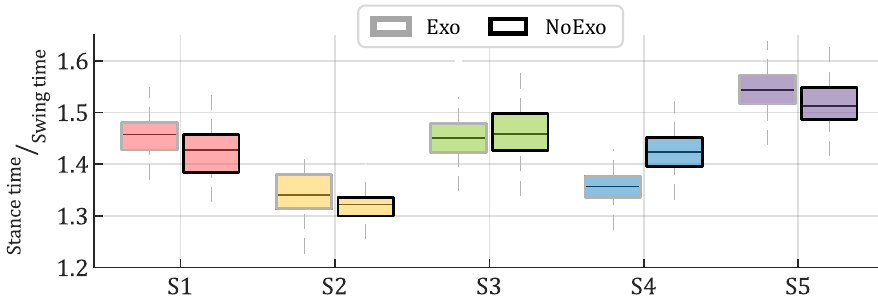}
\caption{The stance-to-swing ratio for each participant, and for each of the two experimental conditions.}
\label{stanceswingt}
\centering
\vspace{-1em}
\end{figure}

\begin{figure}[t]
\includegraphics[width=\linewidth]{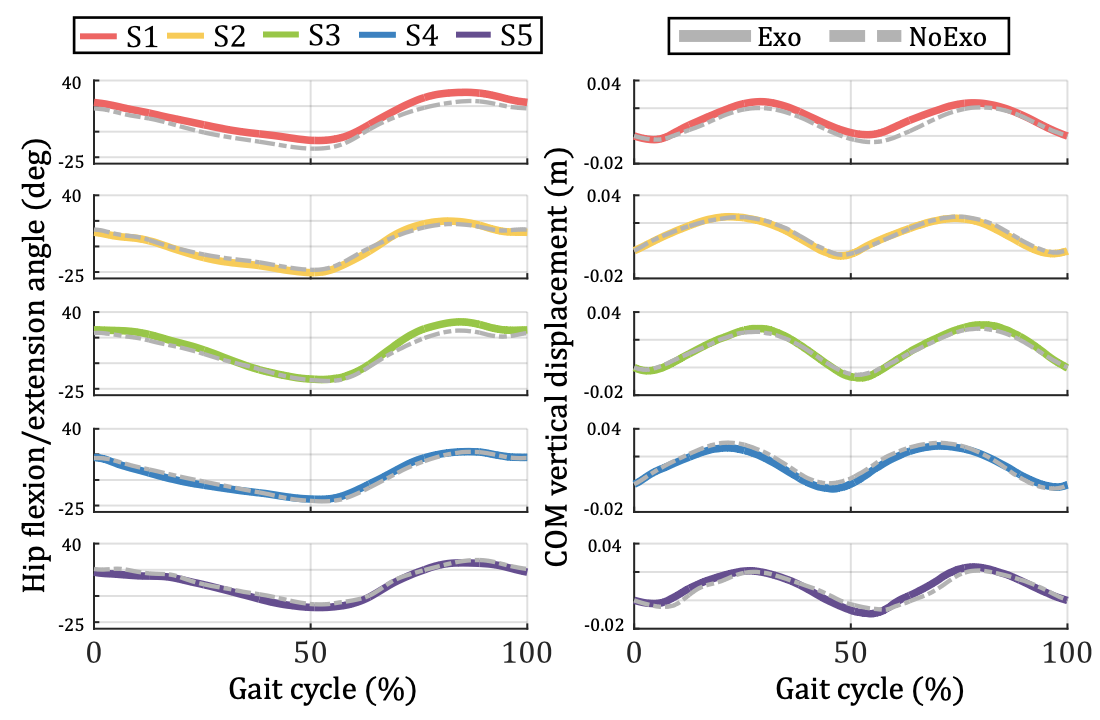}
\caption{Kinematic variables during the assisted and unassisted walking conditions. The center of mass (COM) vertical displacement and the hip flexion/extension angles are shown for each participant to the experiment.}
\label{kinem}
\centering
\vspace{-1.25em}
\end{figure}

\subsubsection{Kinematic data}
During the assisted and unassisted walking conditions, participants adopted a similar walking strategy. This is demonstrated by the consistent stance-to-swing time ratio (Figure \ref{stanceswingt}), and the cadence ratio close to 1.0 for all subjects (Table \ref{tab1}). This finding confirms that our control strategy could assist users without causing forced movements. Conversely, disruptions of wearers' walking kinematics may occur in case of non-synchronized or excessively strong torque profiles \cite{kang_effect_2019}. 

The stance-to-swing time ratio was further analyzed using statistical methods, and a Friedman non-parametric test revealed significant differences between participants. This result can be attributed to the fixed treadmill speed used for all users, which led subjects to adjust their walking strategies accordingly. For instance, we observed that shorter participants adopted a longer stance time to increase stability \cite{winter_human_1995}. These results are consistent with previous research showing that gait phase durations are influenced by anthropometric dimensions \cite{iosa_connection_2016} and deviations from individual preferred walking speed \cite{fukuchi_effects_2019}. On the other hand, within each subject, pairwise post-hoc comparisons indicated no significant difference between assisted and unassisted conditions.   

\begin{figure*}[t]
\centering
\includegraphics[width=\linewidth]{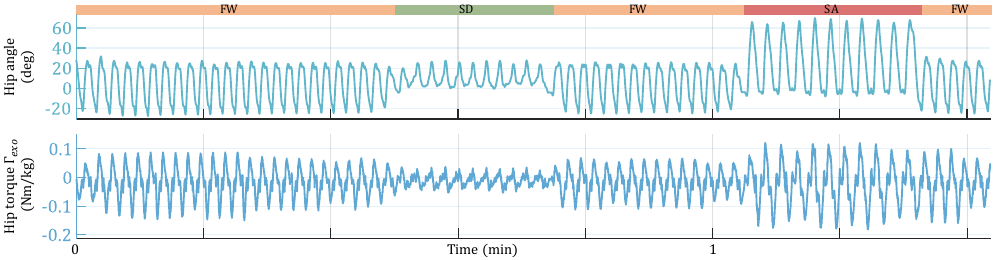}
\caption{Data collected during the varied environment walking trial. The user's hip flexion extension angle is shown, along with the hip assistive torque resulting from our control strategy. Tags: FW = flat walking, SD = stairs descending, SA = stairs ascending.}
\label{kout}
\centering
\vspace{-1.25em}
\end{figure*}

\begin{figure}[t]
\includegraphics[width=\linewidth]{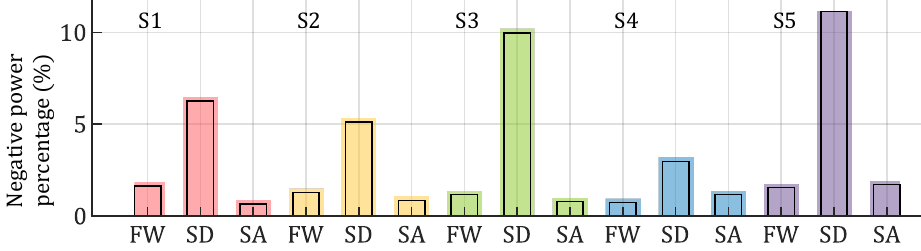}
\caption{Negative power area percentages for each subject in each unstructured environment. Center tags: FW = flat walking, SD = stairs descending, SA = stairs ascending.}
\label{powersOut}
\centering
\vspace{-1.75em}
\end{figure}

Furthermore, for all subjects, the hip joint angle and the vertical displacement of the center of mass (COM) remained unchanged in the two experimental conditions (Figure \ref{kinem}). Avoiding increases or decreases in the COM movement is crucial to avoid higher energetic cost during walking \cite{adamczyk_redirection_2009, wurdeman_reduced_2017, faraji_simple_2018}. This finding also consolidates our results on the decreased metabolic cost measured when walking with the exoskeleton. With our approach, we achieved energetic cost reductions through efficient exoskeleton power transmission and without altering the users' walking kinematics.

\subsubsection{Exoskeleton data - unstructured terrain}
Our control strategy succeeded in assisting users in diverse walking conditions. The exoskeleton provided smooth support during gait transitions, and adapted the assistance to changes in terrain and walking directions (Figure \ref{kout}). 
The assistance strength was dynamically adjusted in real-time through the designed optimization. Our multi-objective Bayesian optimizer remained unaware of the specific walking conditions, and, rather than performing a terrain classification, it relied only on the user's kinematics. 
The optimizer increased the torque strength during stairs ascending, a condition characterized by higher hip power demands \cite{montgomery_contributions_2018, nuckols_mechanics_2020}. Conversely, intermediate assistive torques were applied during flat walking, while the assistance strength was decreased during stair descent. Our findings align with previous research on unstructured walking with hip exoskeletons \cite{lim_delayed_2019, manzoori_adaptive_2024, molinaro_task-agnostic_2024}, reinforcing the importance of modulating assistance strength across different conditions. Achieving this adaptability is essential for delivering seamless exoskeleton support across varied terrains and gait styles transitions \cite{kang_continuous_2020}.

The mechanical power exchanged between the users and the exoskeleton varied depending on the walking scenario (Figure \ref{powersOut}). During flat walking and stairs ascent, the exoskeleton assistance generated low negative power, implying that the exoskeleton's assistive torque was synchronized with the wearers' movements. In contrast, stairs descent exhibited slightly higher negative power (still below $11 \%$ of the total power area), despite the lower assistance strength in this walking condition. These results are consistent with previous findings \cite{lim_delayed_2019}, and highlight the ability of our control framework to provide adaptive and synchronized assistance across varying terrains.


\vspace{-0.5em}
\section{Conclusion}
In this work, we present a novel adaptive and user-dependent control strategy for the hip exoskeleton eWalk. We design the assistive torque as an adaptive virtual negative damping, capable of injecting energy into the human-exoskeleton system while keeping the users in charge of the movements. 
Our approach results in a torque profile that smoothly follows individual gait patters. Additionally, our control strategy reduces the energy expenditure of walking while preserving the natural joint kinematics. When tested in unstructured environments, our multi-objective Bayesian Optimization framework modulates the assistance strength by relying only on the exoskeleton kinematic variables. Notably, the exoskeleton assistance remains synchronized with users' movements across different terrains, as indicated by the low percentages of negative mechanical power transmitted from the device. In future work, we will evaluate the efficacy of the control with a wider population, test other dynamic optimization techniques, and evaluate our control on more unstructured terrains. 

In conclusion, we introduce a simple but powerful strategy to provide partial walking assistance with the hip exoskeleton eWalk. Our control strategy provides individualized and adaptable assistance in various walking scenarios, without the need for extensive modeling or coefficients tuning. Overall, our approach contributes to advancing the development of viable, adaptive, and used-dependent control strategies for hip exoskeletons.


\vspace{-0.5em}
\renewcommand{\baselinestretch}{0.95}  

\bibliographystyle{ieeetr} 
\bibliography{ESA} 

\end{document}